\documentclass[11pt,a4paper]{article}
\usepackage[hyperref]{naaclhlt2019}
\usepackage{times}
\usepackage{latexsym}
\usepackage{graphicx}  
\usepackage{amssymb}
\usepackage{amsmath}
\usepackage{multirow}
\usepackage{color}
\usepackage{hyperref}
\usepackage{fancyhdr}

\usepackage{url}

\aclfinalcopy 

\title{Saliency Learning: Teaching the Model Where to Pay Attention\thanks{\ \ Accepted as a short paper at NAACL HLT 2019}}
\author{
Reza Ghaeini, Xiaoli Z. Fern, Hamed Shahbazi, Prasad Tadepalli\\
School of Electrical Engineering and Computer Science, Oregon State University\\
	1148 Kelley Engineering Center, Corvallis, OR 97331-5501, USA\\
	{\tt \{ghaeinim, xfern, shahbazh, tadepall\}@eecs.oregonstate.edu}\\
 }

\date{}

\begin{document}
\maketitle
\begin{abstract}
Deep learning has emerged as a compelling solution to many NLP tasks with remarkable performances. However, due to their opacity, such models are hard to interpret and trust. Recent work on explaining deep models has introduced approaches to provide insights toward the model's behaviour and predictions, which are helpful for assessing the reliability of the model's predictions. However, such methods do not improve the model's reliability. In this paper, we aim to teach the model to make the right prediction for the right reason by providing explanation training and ensuring the alignment of the model's explanation with the ground truth explanation. Our experimental results on multiple tasks and datasets demonstrate the effectiveness of the proposed method, which produces more reliable predictions while delivering better results compared to traditionally trained models.
\end{abstract}

\section{Introduction}
It is unfortunate that our data is often plagued by meaningless or even harmful statistical biases. When we train a model on such data, it is possible that the classifier focuses on irrelevant biases to achieve high performance on the biased data. Recent studies demonstrate that deep learning models noticeably suffer from this issue~\cite{vqa_biases, squad_biasses, snli_biases}. Due to the black-box nature of deep models and the high dimensionality of their inherent representations, it is difficult to interpret and trust their behaviour and predictions. Recent work on explanation and interpretation has introduced a few approaches \cite{vision_insp, lime, ration, explain_nlp, erasure, explain_mine, anchor} for explanation. Such methods provide insights toward the model's behaviour, which is helpful for detecting biases in our models. However, they do not correct them.  Here, we investigate how to incorporate explanations into the learning process to ensure that our model not only makes correct predictions but also makes them for the right reason.  

Specifically, we propose to train a deep model using both ground truth labels and additional annotations suggesting the desired explanation. The learning is achieved via a novel method called \emph{saliency learning}, which regulates the model's behaviour using saliency to ensure that the most critical factors impacting the model's prediction are aligned with the desired explanation. 

Our work is closely related to \citet{right_for_right}, which also uses the gradient/saliency information to regularize model's behaviour. However, we differ in the following points: 1) \citet{right_for_right} is limited to regularizing model with gradient of the model's input. In contrast, we extend this concept to the intermediate layers of deep models, which is demonstrated to be beneficial based on the experimental results; 2) \citet{right_for_right} considers annotation at the dimension level, which is not appropriate for NLP tasks since the individual dimensions of the word embeddings are not interpretable; 3) most importantly,  \citet{right_for_right} learns from annotations of \emph{irrelevant} parts of the data, whereas we focus on positive annotations identifying parts of the data that contributes positive evidence toward a specific class.  In textual data, it is often unrealistic to annotate a word (even a stop word) to be completely irrelevant. On the other hand, it can be reasonably easy to identify group of words that are positively linked to a class.

We make the following contributions: 1) we propose a new method for teaching the model where to pay attention; 2) we evaluate our method on multiple tasks and datasets and demonstrate that our method achieves more reliable predictions while delivering better results than traditionally trained models; 3) we verify the sensitivity of our saliency-trained model to perturbations introduced on part of the data that contributes to the explanation.

\section{Saliency-based Explanation Learning }
Our goal is to teach the model where to pay attention in order to avoid focusing on meaningless statistical biases in the data. In this work, we focus on positive explanations. In other words, we expect the explanation to highlight information that contributes positively towards the label. For example, if a piece of text contains the mention of a particular event, then the explanation will highlight parts of the text indicating the event, not non-existence of some other events.  This choice is because positive evidence is more natural for human to specify. 

Formally, each training example is a tuple $(X, y, Z)$, where $X = [X_1, X_2, \dots, X_n]$ is the input text (length $n$), $y$ is the ground-truth label, and $Z \in \{0,1\}^{n}$ is the ground-truth explanation as a binary mask indicating whether each word contributes positive evidence toward the label $y$.

Recent studies have shown that the model's predictions can be explained by examining the saliency of the inputs \cite{vision_insp, inp_grad, right_for_right,explain_nlp} as well as other internal elements of the model \cite{explain_mine}. Given an example, for which the model makes a prediction, the saliency of a particular element is computed as the derivative of the model's prediction with respect to that element. 
Saliency provides clues as to where the model is drawing strong evidence to support its prediction. As such, if we constrain the saliency to be aligned with the desired explanation during learning, our model will be coerced to pay attention to the right evidence. 

In computing saliency, we are dealing with high-dimensional data. For example, each word is represented by an embedding of $d$ dimensions. To aggregate the contribution of all dimensions, we consider sum of the gradients of all dimensions as the overall vector/embedding contribution. For the $i$-th word, if $Z[i]=1$, then its vector should have a positive gradient/contribution, otherwise the model would be penalized. To accomplish this, we incorporate a saliency regularization term to the model cost function using hinge loss. Equation~\ref{eq:new:loss} describes our cost function evaluated on a single example $(X,y,Z)$.

\begin{equation}
    \begin{split}
        &\mathcal{C}(\theta, X, y, Z) = \mathcal{L}(\theta, X, y) \\
        &+ \lambda  \sum_{i=1}^n \max \left(0, -Z_{i} \sum_{j=1}^d \frac{\partial f_{\theta}(X,y)}{\partial X_{i,j}}\right)
    \end{split}
    \label{eq:new:loss}
\end{equation}

\noindent where $\mathcal{L}$ is a traditional model cost function (e.g. cross-entropy), $\lambda$ is a hyper parameter, $f$ specifies the model with parameter $\theta$, and $\frac{\partial f}{\partial X_{i,j}}$ represents the saliency of the $j$-th dimension of word $X_i$. The new term in the $\mathcal{C}$ penalizes negative gradient for the marked words in $Z$ (contributory words). 

Since $\mathcal{C}$ is differentiable respect to $\theta$, it can be optimized using existing gradient-based optimization methods. It is important to note that while Equation~\ref{eq:new:loss} only regularizes the saliency of the input layer, the same principle can be applied to the intermediate layers of the model~\cite{explain_mine} by considering the intermediate layer as the input for the later layers.

Note that if $Z = 0$ then $\mathcal{C} = \mathcal{L}$. So, in case of lacking proper annotations for a specific sample or sequence, we can simply use $0$ as its annotation. This property enables our method to be easily used in semi-supervised or active learning settings.

\section{Tasks and Datasets}
To teach the model where to pay attention, we need ground-truth explanation annotation $Z$, which is difficult to come by. As a proof of concept, we modify two well known real tasks (Event Extraction and Cloze-Style Question Answering) to simulate approximate annotations for explanation. Details of the main tasks and datasets could be found in section B of the Appendix. We describe the modified tasks as follows: \\
1) \textbf{Event Extraction:} Given a sentence, the goal is to determine whether the sentence contains an event. Note that event extraction benchmarks contain the annotation of event triggers, which we use to build the annotation $Z$. In particular, the $Z$ value of every word is annotated to be zero unless it belongs to an event trigger. For this task, we consider two well known event extraction datasets, namely ACE 2005 and Rich ERE 2015. \\
2) \textbf{Cloze-Style Question Answering:} Given a sentence and a query with a blank, the goal is to determine whether the sentence contains the correct replacement for the blank. Here, annotation of each word is zero unless it belongs to the gold replacement. For this task, we use two well known cloze-style question answering datasets: Children Book Test Named Entity (CBT-NE) and Common Noun (CBT-CN) \cite{cbt}.

Here, we only consider the simple binary tasks as a first attempt to examine the effectiveness of our method. However, our method is not restricted to binary tasks. In multi-class problems, each class can be treated as the positive class of the binary classification. In such a setting, each class would have its own explanation and annotation $Z$. 

Note that for both tasks if an example is negative, its explanation annotation will be all zero. In other words, for negative examples we have $\mathcal{C} = \mathcal{L}$.

\section{Model}

\begin{figure}[t]
	\centering
	\includegraphics[width=0.49\textwidth]{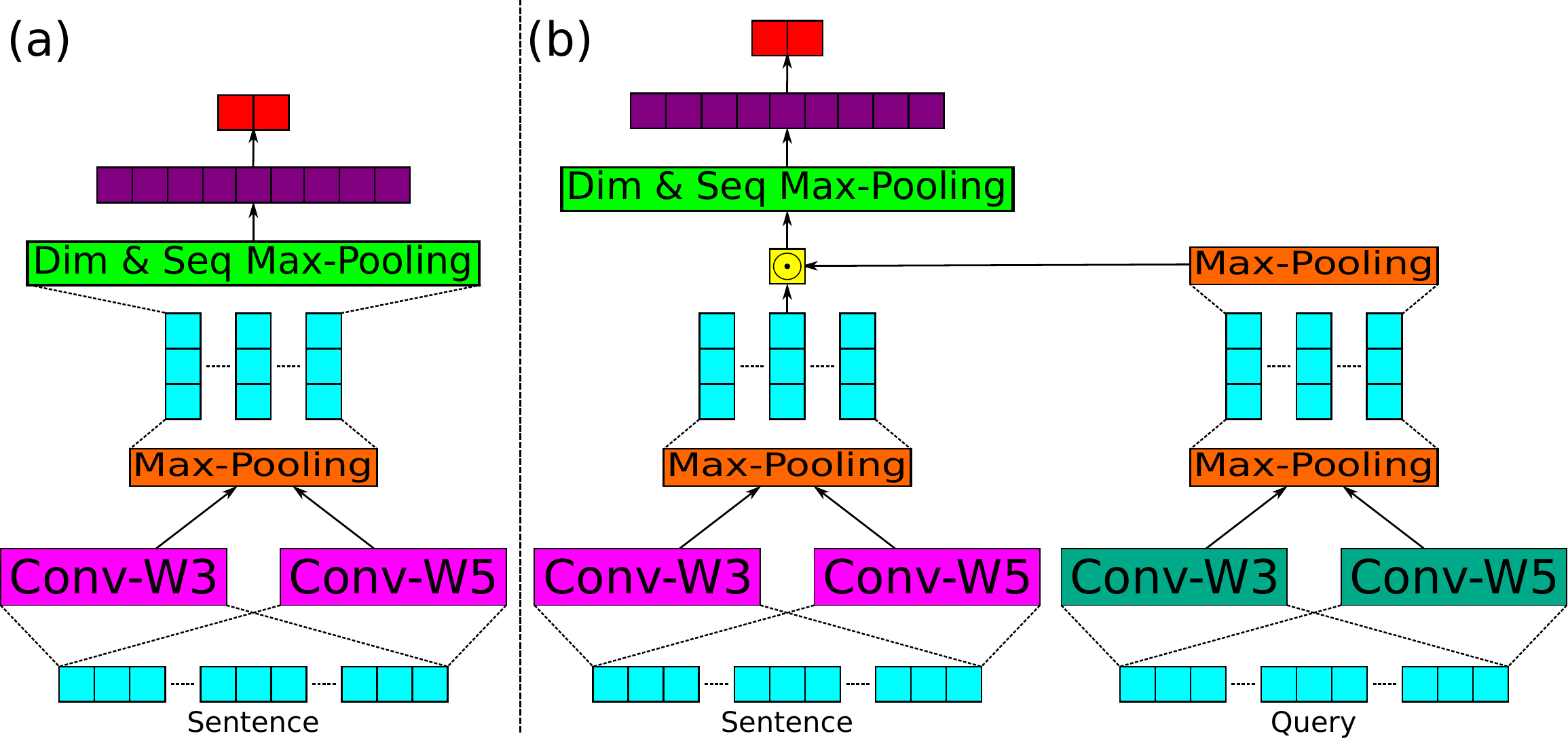}
	\caption{A high-level view of the models used for event extraction (a) and question answering (b). \label{fig:model}}
\end{figure}
We use simple CNN based models to avoid complexity. Figure~\ref{fig:model} illustrates the models used in this paper. Both models have a similar structure. The main difference is that QA has two inputs (sentence and query). We first describe the event extraction model followed by the QA model.

Figure~\ref{fig:model} (a) shows the event extraction model. Given a sentence $W=[w_1, \dots, w_n]$ where $w_i \in \mathbb{R}^{d}$, we first pass the embeddings to two CNNs with feature size of $d$ and window size of $3$ and $5$. Next we apply max-pooling to both CNN outputs. It will give us the representation $I \in \mathbb{R}^{n \times d}$, which we refer to as the \emph{intermediate representation}. Then, we apply sequence-wise and dimension-wise max-poolings to $I$ to capture $D_{seq} \in \mathbb{R}^{d}$ and $D_{dim} \in \mathbb{R}^{n}$ respectively. $D_{dim}$ will be referred as \emph{decision representation}. Finally we pass the concatenation of $D_{seq}$ and $D_{dim}$ to a feed-forward layer for prediction.

Figure~\ref{fig:model} (b) depicts the QA model. The main difference is having \emph{query} as an extra input. To process the query, we use a similar structure to the main model. After CNNs and max-pooling we end up with $Q \in \mathbb{R}^{m \times d}$ where $m$ is the length of query. To obtain a sequence independent vector, we apply another max-pooling to $Q$ resulting in a query representation $q \in \mathbb{R}^{d}$. We follow a similar approach to in event extraction for the given sentence. The only difference is that we apply a dot product between the \emph{intermediate representations} and query representation ($I_i = I_i \odot q$). 

As mentioned previously, we can apply saliency regularization to different levels of the model. In this paper, we apply saliency regularization on the following three levels: 1) Word embeddings ($W$). 2) Intermediate representation ($I$). 3) Decision representation ($D_{dim}$). Note that the aforementioned levels share the same annotation for training. For training details please refer to Section C of the Appendix.

\section{Experiments and Analysis}
\subsection{Performance}
\begin{table}[t]
    \centering
    \begin{tabular}{c|c|c|c|c|c}
        Dataset & S.$^a$ & P.$^b$ & R.$^c$ & F1 & Acc.$^d$ \\ \hline \hline
        \multirow{2}{*}{ACE} & No & 66.0 & \textbf{77.5} & 71.3 & 74.4 \\
        & Yes & \textbf{70.1} & 76.1 & \textbf{73.0} & \textbf{76.9} \\ \hline
        \multirow{2}{*}{ERE} & No & 85.0 & 86.6 & 85.8 & 83.1 \\
        & Yes & \textbf{85.8} & \textbf{87.3} & \textbf{86.6} & \textbf{84.0} \\ \hline \hline
        \multirow{2}{*}{CBT-NE} & No & 55.6 & \textbf{76.3} & 64.3 & 75.5\\
        & Yes & \textbf{57.2} & 74.5 & \textbf{64.7} & \textbf{76.5} \\ \hline
        \multirow{2}{*}{CBT-CN} & No & 47.4 & \textbf{39.0} & 42.8 & 77.3 \\
        & Yes & \textbf{48.3} & 38.9 & \textbf{43.1} & \textbf{77.7} \\ \hline
        \multicolumn{3}{l}{$^a$Saliency Learning.} & \multicolumn{3}{l}{$^b$Precision.}\\
        \multicolumn{3}{l}{$^c$Recall.} & \multicolumn{3}{l}{$^d$Accuracy}\\ \hline
    \end{tabular}
    \caption{Performance of trained models on multiple datasets using traditional method and saliency learning.}
    \label{tab:perform}
\end{table}

Table~\ref{tab:perform} shows the performance of the trained models on ACE, ERE, CBT-NE, and CBT-CN datasets using the aforementioned models with and without saliency learning. The results indicate that using saliency learning yields better accuracy and F1 measure on all four datasets. It is interesting to note that saliency learning consistently helps the models to achieve noticeably higher precision without hurting the F1 measure and accuracy. This observation suggests that saliency learning is effective in providing proper guidance for more accurate predictions -- Note that here we only have guidance for positive prediction. To verify the statistical significance of the observed performance improvement over traditionally trained models without saliency learning, we conducted the one-sided McNemar's test. The obtained p-values are $0.03$, $0.03$, $0.0001$, and $0.04$ for ACE, ERE, CBT-NE, and CBT-CN respectively, indicating that the performance gain by saliency learning is statistically significant.

\subsection{Saliency Accuracy}

In this section, we examine how well does the saliency of the trained model aligns with the annotation. To this end, we define a metric called \emph{saliency accuracy} ($s_{acc}$), which measures what percentage of all positive positions of annotation $Z$ indeed obtain a positive gradient. Formally, $s_{acc} = 100 \frac{\sum_i \delta(Z_i G_i > 0)}{\sum_i Z_i}$ where $G_i$ is the gradient of element $i$ and $\delta$ is the indicator function.

\begin{table}[t]
    \centering
    \begin{tabular}{c|c|c|c|c}
        Dataset & S. & W.$^a$ & I.$^b$ & D.$^c$ \\ \hline \hline
        \multirow{2}{*}{ACE} & No & 61.60 & 66.05 & 63.27 \\
        & Yes & \textbf{99.26} & \textbf{77.92} & \textbf{65.49} \\ \hline
        \multirow{2}{*}{ERE} & No &	51.62 & 56.71 & 44.37 \\
        & Yes & \textbf{99.77} & \textbf{77.45} & \textbf{51.78} \\ \hline \hline
        \multirow{2}{*}{CBT-NE} & No & 52.32 & 65.38 & 68.81 \\
        & Yes & \textbf{98.17} & \textbf{98.34} & \textbf{95.56} \\ \hline
        \multirow{2}{*}{CBT-CN} & No & 47.78 & 53.68 & 45.15 \\
        & Yes & \textbf{99.13} & \textbf{98.94} & \textbf{97.06} \\ \hline
        \multicolumn{5}{l}{$^a$Word Level Saliency Accuracy.}\\
        \multicolumn{5}{l}{$^b$Intermediate Level Saliency Accuracy.} \\
        \multicolumn{5}{l}{$^c$Decision Level Saliency Accuracy.} \\ \hline
    \end{tabular}
    \caption{Saliency accuracy of different layer of our models trained on ACE, ERE, CBT-NE, CBT-CN.}
    \label{tab:saliency:acc}
\end{table}

Table~\ref{tab:saliency:acc} shows the saliency accuracy at different layers of the trained model with and without saliency learning. According to Table~\ref{tab:saliency:acc}, our method achieves a much higher saliency accuracy for all datasets indicating that the learning was indeed effective in aligning the model saliency with the annotation. In other words, important words will have positive contributions in the saliency-trained model, and as such, it learns to focus on the right part(s) of the data. This claim can also be verified by visualizing the saliency, which is provided in the next section.

\subsection{Saliency Visualization}

\begin{table*}[t]
	\centering
	\begin{tabular}{c|l|l||c||c|c}
		id & Baseline Model & Saliency-trained Model & Z & $P_{B}$ & $P_{S}$ \\ \hline
		
		1 & The \colorbox{red!30}{judge} \colorbox{red!40}{at} Hassan's & The judge at \colorbox{red!30}{Hassan}'s & extradition & 1 & 1 \\
		& extradition \colorbox{red!10}{hearing} said   & \colorbox{red!60}{extradition} \colorbox{red!70}{hearing} \colorbox{red!50}{said}  & hearing & & \\
		& \colorbox{red!60}{that} he found the \colorbox{red!70}{French} & that \colorbox{red!20}{he} found the \colorbox{red!40}{French} & said & & \\
		& handwriting report \colorbox{red!20}{very} & \colorbox{red!10}{handwriting} report very & & & \\
		& problematic, very confusing, & problematic, very confusing, & & & \\
		& and with \colorbox{red!50}{suspect} conclusions. & and with suspect conclusions. & & & \\ \hline
		
		2 & Solana said \colorbox{red!40}{the} EU would help  & Solana said the EU would help & attack & 1 & 1 \\
		& \colorbox{red!50}{in} the \colorbox{red!60}{humanitarian} \colorbox{red!70}{crisis} & in the humanitarian \colorbox{red!30}{crisis} & & & \\
		& \colorbox{red!20}{expected} \colorbox{red!30}{to} follow an & \colorbox{red!10}{expected} to \colorbox{red!20}{follow} \colorbox{red!40}{an} & & & \\
		& \colorbox{red!10}{attack} on Iraq. & \colorbox{red!70}{attack} \colorbox{red!60}{on} \colorbox{red!50}{Iraq}. & & & \\ \hline
		
		3 & The trial \colorbox{red!50}{will} start \colorbox{red!70}{on} & The \colorbox{red!70}{trial} will \colorbox{red!10}{start} \colorbox{red!20}{on} & trial & 1 & 1 \\
		& \colorbox{red!60}{March} \colorbox{red!30}{13}, \colorbox{red!10}{the} \colorbox{red!20}{court} \colorbox{red!40}{said}. & \colorbox{red!50}{March} \colorbox{red!60}{13}, \colorbox{red!40}{the} \colorbox{red!30}{court} said. & & & \\ 
		
	\end{tabular}
	\caption{Top 6 salient words visualization of data samples from ACE for the baseline and the saliency-trained models.}
	\label{tab:saliency_sample}
\end{table*}

Here, we visualize the saliency of three positive samples from the ACE dataset for both the traditionally trained (Baseline Model) and the saliency-trained model (saliency-trained Model). Table~\ref{tab:saliency_sample} shows the top 6 salient words (words with highest saliency/gradient) of three positive samples along with their contributory words (annotation $Z$), the baseline model prediction ($P_{B}$), and the saliency-trained model prediction ($P_{S}$). Darker red color indicates more salient words. According to Table~\ref{tab:saliency_sample}, both models correctly predict 1 and the saliency-trained model successfully pays attention to the expected meaningful words while the baseline model pays attention to mostly irrelevant ones. More analyses are provided in section D of the Appendix.

\subsection{Verification}
Up to this point, we show that using saliency learning yields noticeably better precision, F1 measure, accuracy, and saliency accuracy. Here, we aim to verify our claim that saliency learning coerces the model to pay more attention to the critical parts. The annotation $Z$ describes the influential words toward the positive labels. Our hypothesis is that \emph{removing such words would cause more impact on the saliency-trained models} since by training, they should be more sensitive to these words. We measure the impact as the percentage change of the model's true positive rate. This measure is chosen because negative examples do not have any annotated contributory words, and hence we are particularly interested in how removing contributory words of positive examples would impact the model's true positive rate (TPR). 

\begin{table}[t]
    \centering
    \begin{tabular}{c|c|c|c|c}
        Dataset & S. & TPR$_0^a$ & TPR$_1^b$ & $\Delta$TPR$^c$ \\ \hline \hline
        \multirow{2}{*}{ACE} & No & 77.5 & 52.2 & 32.6 \\
        & Yes & 76.1 & 45.0 & \textbf{40.9} \\ \hline
        \multirow{2}{*}{ERE} & No &	86.6 & 73.2 & 15.4 \\
        & Yes & 87.3 & 70.6 & \textbf{19.1} \\ \hline \hline
        \multirow{2}{*}{CBT-NE} & No & 76.3 & 30.2 & 60.4 \\
        & Yes & 74.5 & 28.5 & \textbf{61.8} \\ \hline
        \multirow{2}{*}{CBT-CN} & No & 39.0 & 16.6 & 57.4 \\
        & Yes & 38.9 & 15.4 & \textbf{60.4} \\ \hline
        \multicolumn{5}{l}{$^a$True Positive Rate (before removal).}\\
        \multicolumn{5}{l}{$^b$TPR after removing the critical word(s).} \\
        \multicolumn{5}{l}{$^c$TPR change rate.} \\ \hline
    \end{tabular}
    \caption{True positive rate and true positive rate change of the trained models before and after removing the contributory word(s).}
    \label{tab:positive:swap}
\end{table}

Table~\ref{tab:positive:swap} shows the outcome of the aforementioned experiment, where the last column lists the TPR reduction rates. From the table, we see a consistently higher rate of TPR reduction for saliency-trained models compared to traditionally trained models, suggesting that the saliency-trained models are more sensitive to the perturbation of the contributory word(s) and confirming our hypothesis. 

It is worth noting that we observe less substantial change to the true positive rate for the event task. This is likely due to the fact that we are using trigger words as simulated explanations. While trigger words are clearly related to events, there are often other words in the sentence relating to events but not annotated as trigger words. 
    
\section{Conclusion}
In this paper, we proposed \emph{saliency learning}, a novel approach for teaching a model where to pay attention. We demonstrated the effectiveness of our method on multiple tasks and datasets using simulated explanations. The results show that saliency learning enables us to obtain better precision, F1 measure and accuracy on these tasks and datasets. Further, it produces models whose saliency is more properly aligned with the desired explanation. In other words, \emph{saliency learning} gives us more reliable predictions while delivering better performance than traditionally trained models. Finally, our verification experiments illustrate that the saliency-trained models show higher sensitivity to the removal of contributory words in a positive example. For future work, we will extend our study to examine saliency learning on NLP tasks in an active learning setting where real explanations are requested and provided by a human.

\section*{Acknowledgments}
This material is based upon work supported by the Defense Advanced Research Projects Agency (DARPA) under Contract N66001-17-2-4030.

\bibliography{naaclhlt2019}
\bibliographystyle{acl_natbib}

\newpage

\appendix

\section{Background: Saliency}

The concept of saliency was first introduced in vision for visualizing the spatial support on an image for particular object class \cite{vision_insp}. Considering a deep model prediction as a differentiable model $f$ parameterized by $\theta$ with input $X \in \mathbb{R}^{n \times d}$. Such a model could be described using the Taylor series as follow:

\begin{equation}
f(x) = f(a) + f^{'}(a) (x-a) + \frac{f^{''}(a)}{2!} (x-a)^2 + \dots
\end{equation}

By approximating that a deep model is a linear function, we could use the first order Taylor expansion.

\begin{equation}
f(x) \approx f^{'}(a) x + b
\label{eq:first:taylor}
\end{equation}

\noindent According to Equation~\ref{eq:first:taylor}, the first derivative of the model's prediction respect to its input ($f^{'}(a)$ or $\frac{\partial f}{\partial x}|_{x=a}$) describes the model's behaviour near the input. To make it more clear, bigger derivative/gradient indicates more impact and contribution toward the model's prediction. Consequently, the large-magnitude derivative values determine elements of input that would greatly affect $f(x)$ if changed.

\section{Task and Dataset}
Here, we first describe the main and real Event Extraction and Close-Style Question Answering tasks (before our modification). Next, we provide data statistics of the modified version of ACE, ERE, CBT-NE, and CBT-CN datasets in Table~\ref{tab:data}.

\begin{itemize}
	\item \textbf{Event Extraction:} Given a set of ontologized event types (e.g. Movement, Transaction, Conflict, etc.), the goal of event extraction is to identify the mentions of different events along with their types from natural texts \cite{chen_event,event_nug, walker_event}.
	
	\item \textbf{Cloze-Style Question Answering:} Documents in CBT consist of 20 contiguous sentences from the body of a popular children book and queries are formed by replacing a token from the 21$^{st}$ sentence with a blank. Given a document, a query, and a set of candidates, the goal is to find the correct replacement for blank in the query among the given candidates. To avoid having too many negative examples in our modified datasets, we only consider sentences that contain at least one candidate. To be more clear, each sample from the CBT dataset is split to at most 20 samples -- each sentence of the main sample as long as it contains one of the candidates \cite{epi-reader,as-reader,aoa,ga-reader,cloze_coling}.
	
\end{itemize}

\begin{table}[t]
	\centering
	\begin{tabular}{c||c|c||c|c}
		\multirow{3}{*}{Dataset} & \multicolumn{4}{c}{Sample Count} \\ 
		& \multicolumn{2}{c||}{Train} & \multicolumn{2}{|c}{Test} \\
		& P.$^a$ & N.$^b$ & P. & N. \\ \hline \hline
		ACE & 3.2K & 15K & 293 & 421 \\
		ERE & 3.1K & 4K & 2.7K & 1.91K \\
		CBT-NE & 359K & 1.82M & 8.8K   & 41.1K \\
		CBT-CN & 256K & 2.16M & 5.5K & 44.4K \\
		\hline
		\multicolumn{5}{l}{$^a$ Positive Sample Count} \\
		\multicolumn{5}{l}{$^b$ Negative Sample Count} \\\hline
	\end{tabular}
	\caption{Dataset statistics of the modified tasks and datasets.}
	\label{tab:data}
\end{table}

\begin{figure*}[t]
	\centering
	\includegraphics[width=0.67\textwidth]{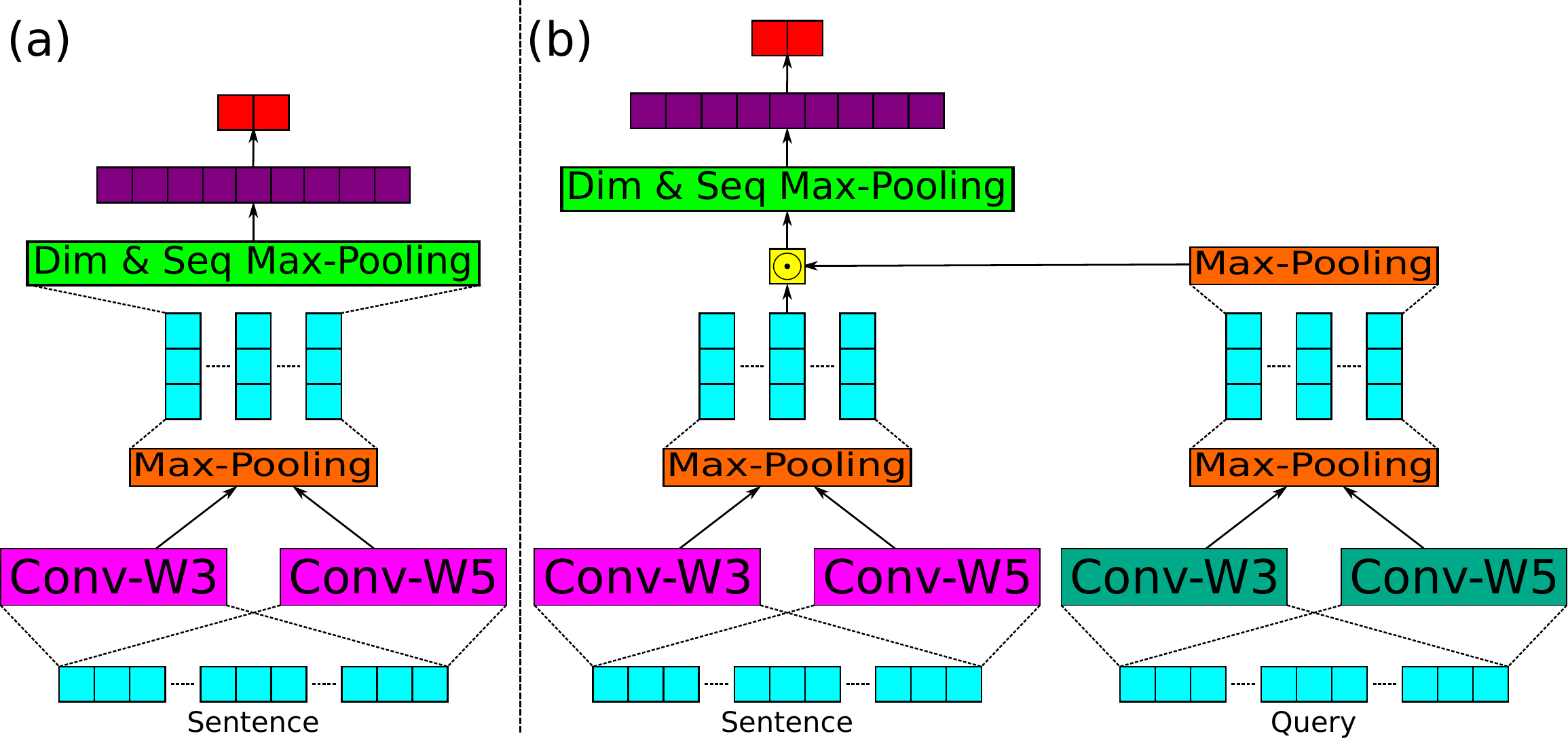}
	\caption{A high-level view of the models used for event extraction(a) and question answering (b). \label{fig:model_big}}
\end{figure*}

\section{Training}
All hyper-parameters are tuned based on the development set. We use pre-trained $300-D$ Glove $840B$ vectors \cite{glove} to initialize our word embedding vectors. All hidden states and feature sizes are $300$ dimensions ($d = 300$). The weights are learned by minimizing the cost function on the training data via Adam optimizer. The initial learning rate is $0.0001$ and $\lambda = 0.5, 0.7, 0.4,$ and $0.35$ for ACE, ERE, CBT-NE, and CBT-CN respectively. To avoid overfitting, we use dropout with a rate of $0.5$ for regularization, which is applied to all feedforward connections. During training, the word embeddings are updated to learn effective representations for each task and dataset. We use a fairly small batch size of $32$ to provide more exploration power to the model. Finally, Equation~\ref{eq:new:loss_all_parts} indicates the the cost function that is used for the training where $W$, $I$, and $D_{dim}$ are the \emph{word embeddings}, \emph{Intermediate representation}, and \emph{Decision representation} respectively.

\begin{equation}
\begin{split}
&\mathcal{C}(\theta, X, y, Z) = \mathcal{L}(\theta, X, y) \\
&+ \lambda  \sum_{i=1}^n \max \left(0, -Z_{i} \sum_{j=1}^d \frac{\partial f_{W}(W,y)}{\partial W_{i,j}}\right) \\
&+ \lambda  \sum_{i=1}^n \max \left(0, -Z_{i} \sum_{j=1}^d \frac{\partial f_{I}(I,y)}{\partial I_{i,j}}\right) \\
&+ \lambda  \sum_{i=1}^n \max \left(0, -Z_{i} \frac{\partial f_{D_{dim}}(D_{dim},y)}{\partial D_{dim, i}}\right) 
\end{split}
\label{eq:new:loss_all_parts}
\end{equation}

\section{Saliency Visualization}

In this section, we empirically analyze the traditionally trained (Baseline Model) and the saliency-trained model (saliency-trained Model) behaviour by observing the saliency of 23 positive samples from ACE and ERE datasets. Tables~\ref{tab:saliency1} and \ref{tab:saliency2} show the top 6 salient words (words with highest saliency/gradient) of positive samples from ACE or ERE dataset along with their contributory word(s) ($Z$), the baseline model prediction ($P_{B}$), and the saliency-trained model prediction ($P_{S}$). Darker red color indicates more salient words. Our observations could be divided into six categories as follow:

\begin{itemize}
	\item Samples 1-7: Both models correctly predict 1 for these samples. The saliency-trained model successfully pays attention to the expected meaningful words while the baseline model pays attention to mostly irrelevant ones.
	
	\item Samples 8-11: Both models correctly predict 1 and pays attention to the contributory words. Yet, we observe lower saliency for important words and higher saliency for irrelevant ones.
	
	\item Samples 12-14: Here, the baseline model fails to pay attention to the contributory words and predicts 0 while the saliency-trained model one successfully pays attention to them and predicts 1.
	
	\item Samples 15-18: Although the models have high saliency for the contributory words, still they could not correctly disambiguate these samples. This observation suggests that having high saliency for important words does not guarantee positive prediction. High saliency for these words indicate their positive contribution toward the positive prediction but still, the model might consider higher probability for negative prediction.
	
	\item Samples 19-21: Here, only the baseline model could correctly predict 1.  However, the baseline model does not pay attention to the contributory words. In other words, the explanation does not support the prediction (unreliable).
	
	\item Samples 22-23: Not always the saliency-trained model could pay proper attention to the contributory words. In these examples, the baseline model has high saliency for contributory words. It is worth noting that when the saliency-trained model does not have high saliency for contributory words, it does not predict 1. Such observation could suggest that the saliency-trained model predictions are more reliable. The aforementioned claim is also verified by consistently obtaining noticeably higher precision for all datasets and tasks (Section 5.1 and Table~\ref{tab:perform} in the paper).
\end{itemize}

\begin{table*}[t]
	\centering
	\resizebox{1.95\columnwidth}{!}{%
	\begin{tabular}{c|l|l||c||c|c}
		id & Baseline Model & Saliency-trained Model & Z & $P_{B}$ & $P_{S}$ \\ \hline
		
		1 & The \colorbox{red!30}{judge} \colorbox{red!40}{at} Hassan's & The judge at \colorbox{red!30}{Hassan}'s & extradition & 1 & 1 \\
		& extradition \colorbox{red!10}{hearing} said   & \colorbox{red!60}{extradition} \colorbox{red!70}{hearing} \colorbox{red!50}{said}  & hearing & & \\
		& \colorbox{red!60}{that} he found the \colorbox{red!70}{French} & that \colorbox{red!20}{he} found the \colorbox{red!40}{French} & said & & \\
		& handwriting report \colorbox{red!20}{very} & \colorbox{red!10}{handwriting} report very & & & \\
		& problematic, very confusing, & problematic, very confusing, & & & \\
		& and with \colorbox{red!50}{suspect} conclusions. & and with suspect conclusions. & & & \\ \hline
		
		2 & Solana said \colorbox{red!40}{the} EU would help  & Solana said the EU would help & attack & 1 & 1 \\
		& \colorbox{red!50}{in} the \colorbox{red!60}{humanitarian} \colorbox{red!70}{crisis} & in the humanitarian \colorbox{red!30}{crisis} & & & \\
		& \colorbox{red!20}{expected} \colorbox{red!30}{to} follow an & \colorbox{red!10}{expected} to \colorbox{red!20}{follow} \colorbox{red!40}{an} & & & \\
		& \colorbox{red!10}{attack} on Iraq. & \colorbox{red!70}{attack} \colorbox{red!60}{on} \colorbox{red!50}{Iraq}. & & & \\ \hline
		
		3 & The trial \colorbox{red!50}{will} start \colorbox{red!70}{on} & The \colorbox{red!70}{trial} will \colorbox{red!10}{start} \colorbox{red!20}{on} & trial & 1 & 1 \\
		& \colorbox{red!60}{March} \colorbox{red!30}{13}, \colorbox{red!10}{the} \colorbox{red!20}{court} \colorbox{red!40}{said}. & \colorbox{red!50}{March} \colorbox{red!60}{13}, \colorbox{red!40}{the} \colorbox{red!30}{court} said. & & & \\ \hline
		
		4 & \colorbox{red!50}{India}'s has been reeling & \colorbox{red!30}{India}'s \colorbox{red!10}{has} been reeling & killed & 1 & 1 \\
		& \colorbox{red!20}{under} a \colorbox{red!30}{heatwave} since & under a heatwave since & & & \\
		& \colorbox{red!40}{mid-May} which \colorbox{red!60}{has} & \colorbox{red!20}{mid-May} which \colorbox{red!60}{has}  & & & \\ 
		& \colorbox{red!10}{killed} \colorbox{red!70}{1,403} people. & \colorbox{red!70}{killed} \colorbox{red!50}{1,403} \colorbox{red!40}{people}. & & & \\ \hline
		
		5 & Retired General Electric Co. & Retired General Electric Co. & Retired & 1 & 1 \\
		& \colorbox{red!20}{Chairman} Jack \colorbox{red!10}{Welch} is & Chairman Jack Welch is & divorce & & \\
		& seeking work-related & seeking work-related & & & \\
		& \colorbox{red!30}{documents} of \colorbox{red!40}{his} \colorbox{red!60}{estranged} & documents of his \colorbox{red!10}{estranged} & & & \\
		& wife in his \colorbox{red!50}{high-stakes} & \colorbox{red!30}{wife} \colorbox{red!50}{in} \colorbox{red!60}{his} \colorbox{red!40}{high-stakes} & & & \\
		& divorce \colorbox{red!70}{case}. & \colorbox{red!70}{divorce} \colorbox{red!20}{case}. & & & \\ \hline
		
		6 & The following year, he \colorbox{red!50}{was} & The following year, \colorbox{red!50}{he} \colorbox{red!40}{was} & acquitted & 1 & 1 \\
		& acquitted \colorbox{red!20}{in} the \colorbox{red!60}{Guatemala} & \colorbox{red!30}{acquitted} \colorbox{red!60}{in} the Guatemala & case & & \\
		& case, \colorbox{red!40}{but} \colorbox{red!70}{the} \colorbox{red!30}{U.S.} continued & \colorbox{red!70}{case}, but the U.S. \colorbox{red!10}{continued} & & & \\
		& to \colorbox{red!10}{push} for his prosecution. & to push for his \colorbox{red!20}{prosecution}. & & & \\ \hline
		
		7 & In 2011, \colorbox{red!50}{a} Spanish \colorbox{red!60}{National} & In 2011, \colorbox{red!30}{a} \colorbox{red!10}{Spanish} \colorbox{red!20}{National} & issued & 1 & 1 \\
		& Court \colorbox{red!70}{judge} issued arrest & \colorbox{red!50}{Court} judge \colorbox{red!70}{issued} \colorbox{red!60}{arrest} & slaying & & \\
		& warrants for 20 \colorbox{red!30}{men}, & \colorbox{red!40}{warrants} for 20 men, & arrest & & \\
		& including Montano,suspected & including Montano,suspected & & & \\
		& \colorbox{red!40}{of} participating in the & of participating in the & & & \\
		& slaying \colorbox{red!10}{of} \colorbox{red!20}{the} priests. & slaying of the priests. & & & \\ \hline
		
		8 & Slobodan Milosevic's wife will & Slobodan Milosevic's wife will & trial & 1 & 1 \\
		& go \colorbox{red!50}{on} \colorbox{red!70}{trial} \colorbox{red!10}{next} week \colorbox{red!20}{on} & go on \colorbox{red!70}{trial} \colorbox{red!20}{next} week on & charges & & \\
		& charges of mismanaging \colorbox{red!30}{state} & \colorbox{red!10}{charges} of mismanaging state & former & & \\
		& property during the \colorbox{red!40}{former} & property during the former & & & \\
		& president's rule, \colorbox{red!60}{a} court said & \colorbox{red!30}{president}'s rule, a \colorbox{red!40}{court} \colorbox{red!60}{said} & & & \\ 
		& Thursday. & \colorbox{red!50}{Thursday}. & & & \\ \hline
		
		9 & \colorbox{red!50}{Iraqis} \colorbox{red!20}{mostly} \colorbox{red!40}{fought} \colorbox{red!70}{back} & \colorbox{red!30}{Iraqis} \colorbox{red!60}{mostly} \colorbox{red!70}{fought} \colorbox{red!20}{back} & fought & 1 & 1  \\
		& with \colorbox{red!60}{small} arms, pistols, & \colorbox{red!50}{with} small arms, pistols, & & & \\
		&  \colorbox{red!30}{machine} guns and &  machine \colorbox{red!40}{guns} \colorbox{red!10}{and} & & & \\ 
		& rocket-propelled \colorbox{red!10}{grenades}. & rocket-propelled grenades. & & & \\ \hline 
		
		10 & \colorbox{red!10}{But} \colorbox{red!20}{the} Saint Petersburg & But the Saint Petersburg & summit & 1 & 1 \\
		& \colorbox{red!30}{summit} \colorbox{red!60}{ended} without \colorbox{red!50}{any} & \colorbox{red!70}{summit} \colorbox{red!60}{ended} \colorbox{red!10}{without} \colorbox{red!30}{any}  & & & \\
		& \colorbox{red!70}{formal} declaration on \colorbox{red!40}{Iraq}. & formal \colorbox{red!20}{declaration} \colorbox{red!40}{on} \colorbox{red!50}{Iraq}. & & & \\
		
	\end{tabular}%
	}
	\caption{Top 6 salient words visualization of samples from ACE and ERE for the baseline and the saliency-trained models.}
	\label{tab:saliency1}
\end{table*}

\begin{table*}[t]
	\centering
	\resizebox{1.98\columnwidth}{!}{%
	\begin{tabular}{c|l|l||c||c|c}
		id & Baseline Model & Saliency-trained Model & Z & $P_{B}$ & $P_{S}$ \\ \hline 
		
		11 & He \colorbox{red!50}{will} then stay on \colorbox{red!70}{for} \colorbox{red!30}{a} & He will then \colorbox{red!20}{stay} on for \colorbox{red!30}{a} & heading & 1 & 1 \\
		& regional summit before & \colorbox{red!60}{regional} \colorbox{red!70}{summit} \colorbox{red!40}{before} & summit & & \\
		& \colorbox{red!60}{heading} to Saint Petersburg & \colorbox{red!50}{heading} \colorbox{red!10}{to} Saint Petersburg  & & & \\
		& \colorbox{red!10}{for} celebrations marking the & for celebrations marking the & & & \\ 
		& \colorbox{red!20}{300th} anniversary of the & 300th anniversary of the & & & \\ 
		& city's \colorbox{red!40}{founding}. & city's founding. & & & \\ \hline
		
		12 & \colorbox{red!30}{From} \colorbox{red!70}{greatest} \colorbox{red!10}{moment} of & From greatest moment of & divorce & 0 & 1 \\ 
		& his \colorbox{red!40}{life} \colorbox{red!60}{to} divorce \colorbox{red!20}{in} 3 & \colorbox{red!10}{his} \colorbox{red!40}{life} \colorbox{red!50}{to} \colorbox{red!70}{divorce} \colorbox{red!60}{in} \colorbox{red!30}{3} & & & \\
		& \colorbox{red!50}{years} or less. & \colorbox{red!20}{years} or less. & & & \\ \hline
		
		13 & The \colorbox{red!10}{state}’s \colorbox{red!40}{execution} \colorbox{red!50}{record} & The \colorbox{red!30}{state}’s \colorbox{red!70}{execution} \colorbox{red!60}{record} & execution & 0 & 1 \\
		& \colorbox{red!60}{has} \colorbox{red!70}{often} \colorbox{red!20}{been} \colorbox{red!30}{criticized}. & \colorbox{red!50}{has} \colorbox{red!20}{often} \colorbox{red!10}{been} \colorbox{red!40}{criticized}. & & & \\ \hline
		
		14 & The student, who \colorbox{red!20}{was} 18 at & The student, who was \colorbox{red!20}{18} at & testified & 0 & 1 \\
		& \colorbox{red!70}{the} \colorbox{red!60}{time} \colorbox{red!50}{of} \colorbox{red!40}{the} alleged & the time of the \colorbox{red!60}{alleged} & & & \\
		& sexual relationship, testified & \colorbox{red!70}{sexual} \colorbox{red!50}{relationship}, \colorbox{red!40}{testified} & & & \\
		& \colorbox{red!30}{under} a \colorbox{red!10}{pseudonym}. & \colorbox{red!10}{under} \colorbox{red!30}{a} pseudonym. & & & \\ \hline
		
		15 & \colorbox{red!70}{U.S.} \colorbox{red!40}{aircraft} \colorbox{red!30}{bombed} \colorbox{red!60}{Iraqi}  & \colorbox{red!20}{U.S.} \colorbox{red!60}{aircraft} \colorbox{red!70}{bombed} \colorbox{red!50}{Iraqi}  & bombed & 0 & 0 \\
		& \colorbox{red!20}{tanks} \colorbox{red!50}{holding} bridges close & \colorbox{red!40}{tanks} \colorbox{red!30}{holding} \colorbox{red!10}{bridges} close & & & \\
		& to the \colorbox{red!10}{city}. & to the city. & & & \\ \hline
		
		16 & \colorbox{red!10}{However}, no blasphemy  & However, no blasphemy & executed & 0 & 0 \\ 
		& \colorbox{red!40}{convict} has \colorbox{red!20}{ever} \colorbox{red!30}{been} & \colorbox{red!20}{convict} \colorbox{red!10}{has} ever \colorbox{red!30}{been} & & & \\
		& \colorbox{red!50}{executed} in \colorbox{red!60}{the} \colorbox{red!70}{country}. & \colorbox{red!60}{executed} \colorbox{red!50}{in} \colorbox{red!40}{the} \colorbox{red!70}{country}. & & & \\ \hline
		
		17 & \colorbox{red!60}{Gul}'s \colorbox{red!30}{resignation} \colorbox{red!70}{had} & \colorbox{red!20}{Gul}'s \colorbox{red!70}{resignation} \colorbox{red!50}{had} & resignation & 0 & 0 \\
		& \colorbox{red!40}{been} \colorbox{red!20}{long} \colorbox{red!50}{expected}. & \colorbox{red!60}{been} \colorbox{red!30}{long} \colorbox{red!40}{expected}. & & & \\ \hline
		
		18 & aside \colorbox{red!50}{from} \colorbox{red!20}{purchasing} & aside \colorbox{red!30}{from} \colorbox{red!70}{purchasing} & purchasing & 0 & 0 \\ 
		& alcohol, \colorbox{red!70}{what} \colorbox{red!10}{rights} & \colorbox{red!60}{alcohol}, \colorbox{red!40}{what} \colorbox{red!10}{rights} & & & \\ 
		& \colorbox{red!60}{don't} 18 \colorbox{red!40}{year} \colorbox{red!30}{olds} have? & \colorbox{red!50}{don't} \colorbox{red!20}{18} year olds have? & & & \\ \hline
		
		19 & \colorbox{red!10}{He} \colorbox{red!50}{also} ordered \colorbox{red!30}{him} to & He also \colorbox{red!60}{ordered} him to & ordered & 1 & 0 \\
		& have \colorbox{red!70}{no} \colorbox{red!20}{contact} \colorbox{red!40}{with} & \colorbox{red!10}{have} \colorbox{red!30}{no} \colorbox{red!50}{contact} \colorbox{red!70}{with} & contact & & \\ 
		& \colorbox{red!60}{Shannon} Molden. &  \colorbox{red!40}{Shannon} \colorbox{red!20}{Molden}. & & & \\  \hline
		
		20 & This \colorbox{red!20}{means} \colorbox{red!10}{your} \colorbox{red!60}{account} is & This \colorbox{red!10}{means} your \colorbox{red!20}{account} is & wrote & 1 & 0 \\ 
		& \colorbox{red!40}{once} \colorbox{red!30}{again} \colorbox{red!50}{active} and & once again active and & & & \\
		& operational, Riaño wrote & \colorbox{red!40}{operational}, \colorbox{red!50}{Riaño} \colorbox{red!70}{wrote}  & & & \\ 
		& \colorbox{red!70}{Colombia} Reports. & \colorbox{red!60}{Colombia} \colorbox{red!30}{Reports}. & & & \\ \hline
		
		21 & I \colorbox{red!70}{am} \colorbox{red!50}{a} Christian as \colorbox{red!40}{is} & I am \colorbox{red!20}{a} \colorbox{red!40}{Christian} as is & divorced & 1 & 0 \\
		& \colorbox{red!30}{my} \colorbox{red!10}{ex} husband \colorbox{red!60}{yet} & my \colorbox{red!30}{ex} \colorbox{red!60}{husband} yet & ex & & \\ 
		& \colorbox{red!20}{we} are divorced. & \colorbox{red!10}{we} \colorbox{red!70}{are} \colorbox{red!50}{divorced}. & & & \\ \hline
		
		22 & Taylor acknowledged in his & Taylor acknowledged \colorbox{red!20}{in} his & testimony & 1 & 0 \\
		& \colorbox{red!70}{testimony} that he \colorbox{red!50}{ran} up & \colorbox{red!10}{testimony} that he \colorbox{red!40}{ran} up & followed & & \\
		& \colorbox{red!60}{toward} \colorbox{red!40}{the} pulpit with \colorbox{red!30}{a} & toward the pulpit \colorbox{red!70}{with} \colorbox{red!50}{a} & ran & & \\
		& \colorbox{red!20}{large} group and \colorbox{red!10}{followed}  & \colorbox{red!30}{large} \colorbox{red!60}{group} and followed & & & \\
		& the men outside. & the men outside. & & & \\ \hline
		
		23 & The \colorbox{red!60}{note} admonished \colorbox{red!70}{Jasper} & The \colorbox{red!10}{note} \colorbox{red!50}{admonished} \colorbox{red!60}{Jasper} & note & 0 & 0 \\
		& \colorbox{red!40}{Molden}, \colorbox{red!50}{and} \colorbox{red!20}{his} then-fiancée, & \colorbox{red!70}{Molden}, and \colorbox{red!20}{his} \colorbox{red!40}{then-fiancée}, & & & \\
		& \colorbox{red!30}{Shannon} \colorbox{red!10}{Molden}. & \colorbox{red!30}{Shannon} Molden. & & & \\ 
		
	\end{tabular}%
	}
	\caption{Top 6 salient words visualization of samples from ACE and ERE for the baseline and the saliency-trained models.}
	\label{tab:saliency2}
\end{table*}

\end{document}